\documentclass{article} 
\usepackage{iclr2020_conference,times}

\usepackage{amsmath,amsfonts,bm}

\def\eqref#1{equation~\ref{#1}}

\def\1{\bm{1}}

\DeclareMathAlphabet{\mathsfit}{\encodingdefault}{\sfdefault}{m}{sl}
\SetMathAlphabet{\mathsfit}{bold}{\encodingdefault}{\sfdefault}{bx}{n}

\usepackage{hyperref}
\usepackage{url}
\usepackage{times}
\usepackage{latexsym}
\usepackage{tabularx}
\usepackage{cleveref}
\usepackage{makecell}
\usepackage{float}
\usepackage{graphicx}
\usepackage[toc,page]{appendix}
\Crefname{appsec}{Appendix}{Appendices}

\title{In-training Matrix Factorization for Parameter-frugal Neural Machine Translation}

\author{\centerline{Zachary Kaden \qquad Teven Le Scao \qquad Raphael Olivier\thanks{Corresponding author}} \\
\centerline{Language Technologies Institute} \\
\centerline{Carnegie Mellon University} \\
\centerline{Pittsburgh, PA} \\
\centerline{{\tt \{zkaden, tlescao, rolivier\}@cs.cmu.edu}}}

\iclrfinalcopy 
\begin{document}

\maketitle

\begin{abstract}
  In this paper, we propose the use of in-training matrix factorization to reduce the model size for neural machine translation. Using in-training matrix factorization, parameter matrices may be decomposed into the products of smaller matrices, which can compress large machine translation architectures by vastly reducing the number of learnable parameters. We apply in-training matrix factorization to different layers of standard neural architectures and show that in-training factorization is capable of reducing nearly 50\% of learnable parameters without any associated loss in BLEU score. Further, we find that in-training matrix factorization is especially powerful on embedding layers, providing a simple and effective method to curtail the number of parameters with minimal impact on model performance, and, at times, an increase in performance.
\end{abstract}

\section{Introduction}
While neural models for machine translation have realized considerable breakthroughs in recent years and are now state-of-the-art in many contexts, they are frequently expensive in resources, owing to their large number of parameters. In a context of democratization of deep learning tools, having smaller models that can be learned and/or applied offline on small devices would have immediate and important applications, for example for privacy reasons. However, it is often necessary to leverage deep networks with large layers in order to capture abstract and complex patterns and interactions over the data. We posit that the need for such large parameter matrices is not because they are essential to the representational power of the network. Rather, our hypothesis is that having many parameters can help neural architectures overcome limitations introduced with approximate optimization methods, noisy data supervision, and sub-optimal model architectures. Moreover, recent work has suggested many parameters in large neural architectures are largely superfluous, serving solely to accommodate the stochastic nature of modern machine learning optimization algorithms \citep{lottery_ticket}.

Motivated by those hypotheses, we study the application of in-training matrix factorization to neural machine translation. Traditional matrix factorization methods are a simple but powerful technique that has been used after training in other deep learning systems, such as \citet{6638949} for speech recognition and \citet{DBLP:journals/corr/KimPYCYS15} for computer vision. In contrast to traditional matrix factorization techniques, in-training matrix factorization reduces the number of learnable parameters at training time, lessening the need for computational resources.

The main contributions of this work are:
\begin{enumerate}
    \item We formally define in-training matrix factorization and present a technique to utilize matrix factorization during training time.
    \item We conduct sets of experiments on two standard neural machine translation architectures: the LSTM encoder-decoder and the transformer network.
    \item We show that in-training factorization can decrease a model's size by half with no impact on performance, and at times, can improve model performance 
\end{enumerate}

\section{Approach}\label{sec : approach}

\subsection{Traditional post-training matrix factorization}
Traditionally, matrix factorization has been used as a compression method for a pretrained model. In this case one must obtain a low-rank approximation of a matrix $M$. A solution is to follow the Eckart-Young-Mirsky theorem \citep{Eckart1936} and use singular-value decomposition (SVD).

If $M=UDV^\top$ with $U \in \mathbb{R}^{n,n}$, $V \in \mathbb{R}^{m,m}$ and $D = Diag(\lambda_1, ... , \lambda_m) \in \mathbb{R}^{n,m}$ with $(\lambda_i)$ in decreasing order, then by keeping the top $p$ diagonal elements of $D$ and the corresponding matrix blocks in $U$ and $V$ we get the best $p$-rank approximation of $M$ for the $L_2$ (Frobenius) norm. 

In practice, because this factorization disturbs the computation graph, the result of this factorization does not always reproduce the non-factorized performance. Thus, practitioners generally must, additionally, refine their compressed models by relearning the pretrained and factorized parameter matrices.

This form of post-training matrix factorization suffers from two core limitations. First, the original uncompressed model must be learned. This results in increased training time, computational resources, and cost to model development. Second, the factorized model must be retrained before it can be deployed, which, again, increases  training time, computational resources, and cost.

\subsection{In-training factorization}
\label{sec: in-training}
\begin{figure}
    \centering
    \includegraphics[width=\textwidth]{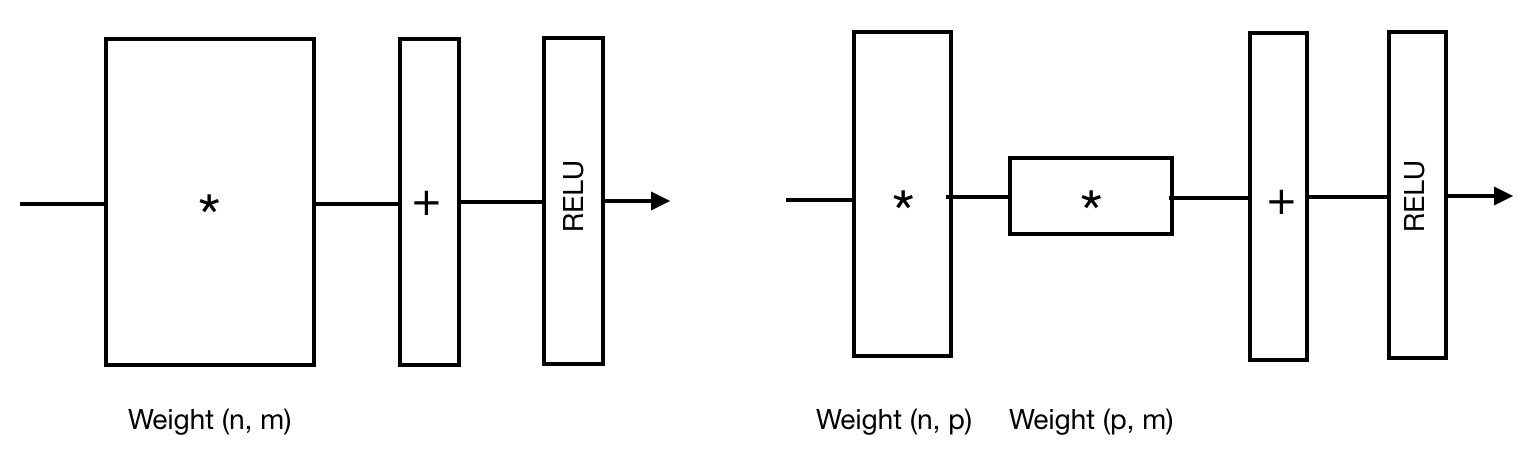}
    \caption{A diagram of in-training factorization. A weight matrix is replaced by the product of two weight matrices, followed by the bias and activation.}
    \label{fig : in-training-diagram}
\end{figure}

It is, however, even easier to initialize the neural network in factorized form, and perform training from there. The idea of in-training matrix factorization is straightforward: wherever the computation graph involves multiplication by a $(n,m)$ parameter matrix $M$, we replace $M$ by the product of two matrices of sizes $(n,p)$ and $(p,m) $, where $p << min(n,m)$. A diagram of in-training factorization is provided in \Cref{fig : in-training-diagram}. Algebraically speaking, this is a $p$-rank approximation of $M$. We refer to $n, m,$ and $p$ as the \textit{left}, \textit{right} and \textit{inner} sizes, respectively, of the in-training factorized matrix.

In terms of implementation, we replace one linear layer with two, with a single output bias and activation after the second layer. Thus, this method replaces $ (nm) $-many parameters with $p * (n + m) $-many parameters. Hence, when $p<\frac{nm}{n+m}$, we reduce the number of parameters. For square matrices where $n = m $, the limit is $p<\frac{m}{2} $. In our experiments, we always set $p \leq \frac{min(n, m)}{2} $.

Even though it may seem that the structure of the network has not changed, as multiplying the two matrices would result in a standard layer with a lower-rank weight matrix, this form seems very adapted to optimization constraints. Applied on embedding and projection layers especially, this simple technique is surprisingly effective, and the change is almost transparent. It can also be combined with weight tying \citep{press2016} without difficulty.

\section{Experiments}\label{sec : experiments}

\subsection{Dataset}
Unless specified otherwise, we conduct experiments using the English-to-German task of the IWSLT 2014 dataset. It consists of transcripts of TED talks translated into several languages, with sentence-to-sentence correspondences. The training, validation, and test sets contain 153326, 6969, and 6750 sentences, respectively. We perform additional experiments using the English-to-Portuguese and English-to-Turkish language pairs from the same dataset.

\subsection{Models}
We run separate sets of experiments on two standard neural machine translation architectures: an LSTM encoder-decoder and a transformer network. 

\subsubsection{LSTM encoder-decoder} This architecture mostly follows \citet{luong2015}. We use a 3-layer bidirectional encoder with hidden size 256 in each direction, and a 3-layer decoder with size 512. The embedding size is 256 and the embedding layer of the decoder is tied with the projection layer, following \citet{press2016}. Beam search is used for inference. The vocabulary size is 30k for the source, 46k for the target.

\subsubsection{Transformer} This model \citep{vaswani2017} makes use of multi-head attention and self attention rather than a recurrent architecture. We adapt the implementation provided by \citet{opennmt}, including the learning rate schedule, with smaller parameters, which are provided in \Cref{tab : transformer_by_language}. We accumulate 8 mini-batches of 12 to 16 sentences between gradient updates to arrive at the total batch sizes we're reporting. The reported experiments do not tie the embedding and projection layers -- which when tried did not yield satisfying results. This model uses subwords and has 15k subword vocabulary for source and target language, respectively.

\subsection{Methodology}
For a given model, we run in-training factorization experiments with different inner sizes $p$, on different layers. We add comparisons with non-factorized models of equivalent size $p$. We compare our method with a compression baseline proposed by \citet{see2016}, based solely on the pruning of small elements for compression purposes. However, our method has the crucial advantage of being compatible with modern GPU computation, while pruning creates sparse matrices that do not benefit as much from GPU acceleration. Additionally, we run post-training factorization with a variety of schemes and compare results with pruning and in-training factorization.

We hold batch sizes constant when training across the various compression paradigms. We do so in order to compare compression methods in training environments that are as similar as possible. We note, however, that in-training factorization provides the ability to increase the batch size by compressing the model during training. It has been shown that larger batch sizes can decrease training time and increase performance, especially on the transformer model \citep{DBLP:journals/corr/abs-1804-00247}. Larger batch sizes can be simulated by accumulating mini-batch gradients before performing the gradient update, which aids in performance; however, accumulating gradients does not efficiently parallelize operations as can a GPU with larger batches.

\section{Results and Analysis}\label{sec : results}

\subsection{In-training factorization}
\Cref{tab : lstm} provides results and comparisons for in-training factorization with the LSTM encoder-decoder model. Tables \ref{tab : transformer} and \ref{tab : transformer post-training} provide results and comparisons for the transformer model. We observe a number of findings from the results.

\subsubsection{Parameter reduction}

We observe that in-training factorization can bring a significant reduction of the number of parameters with minimal impact to performance. With the LSTM we can reduce model size by 48\% with no drop in BLEU score, and we can reduce model size by 56\% and lose less than 0.1 BLEU points. With the transformer model, we can reduce model size by 27\% with no drop in BLEU score, and we can reduce the model size by 51\% with only a 0.4 drop in BLEU score.  By comparison, the same size reductions through pruning decreases the BLEU score by 1.8 BLEU points on the LSTM model and 2.8 BLEU points on the transformer model.

\begin{table*}
\centering
\caption{ \label{tab : lstm}Results of factorization methods for the LSTM encoder-decoder model. ``Pruned" refers to a model where the smallest x\% of parameters have been pruned. ``In-training" refers to the in-training factorization method described in \Cref{sec: in-training} performed on the embedding/projection layer.}
\vspace*{.5cm}
\begin{tabular}{|l|l|l|l|}
\hline
\textbf{Compression method} & \textbf{Size reduction} & \textbf{BLEU} & \textbf{BLEU non-factorized}
\\ \Xhline{3\arrayrulewidth}
None, embedding=256 (baseline) & 0\% & - &  26.70
\\ \hline
Pruned, embedding=256  & -40\%  & 26.61  & -
\\ \hline
Pruned, embedding=256  & -56\%  & 24.91  & -
\\ \Xhline{2\arrayrulewidth}
In-training, inner size=128 &  -32\% & 26.0 & 26.47
\\ \hline
In-training, inner size=64 & -48\% & 26.75 & 25.34
\\ \hline
In-training, inner size=32 & -56\% & 26.63 & 22.69
\\ \hline
In-training, inner size=16 & -60\% & 24.72 & 11.74
\\ \hline
\end{tabular}
\end{table*}

\begin{table*}
\centering
\caption{ \label{tab : transformer}Results of in-training factorization for the transformer architecture. ``Pruned" refers to a model where the smallest x\% of parameters have been pruned. ``In-training (embed)" refers to in-training factorization performed on the embedding and projection layers. ``In-training (+feed-forward)" refers to in-training factorization performed on the embedding, projection, and feed forward layers. ``In-training (+attention)" refers to in-training factorization performed on the embedding, projection, feed forward, and attention layers.}
\vspace*{.5cm}
\resizebox{\textwidth}{!}{
\begin{tabular}{|l|l|l|l|}
\hline
\textbf{Compression method} & \textbf{Size reduction} & \textbf{BLEU} & \textbf{BLEU non-factorized}
\\ \Xhline{3\arrayrulewidth}
None, embedding=512 (baseline) & 0\% & - & 25.95
\\ \hline
Pruned, embedding=512  & -27\%  & 25.67 & -
\\ \hline
Pruned, embedding=512  & -51\%  & 23.15  & -
\\ \Xhline{2\arrayrulewidth}
In-training (embed), inner size=256 & -19\% & 26.22 & 25.65
\\ \hline
In-training (embed), inner size=128 & -27\%  & 26.11 & 23.88
\\ \hline
In-training (embed), inner size=64 & -51\% & 25.55 & 20.01
\\ \hline
In-training (embed), inner size=32 & -55\% & 24.75 & 12.42
\\ \hline
In-training (+feed-forward), inner size=256 & -32\% & 25.58 & 25.65
\\ \hline
In-training (+feed-forward), inner size=128 & -54\% & 25.39 & 23.88
\\ \hline
In-training (+feed-forward), inner size=64 & -64\% & 25.24  & 20.01
\\ \hline
In-training (+feed-forward), inner size=32 & -70\% & 22.43  & 12.42
\\ \hline
In-training (+attention), inner size=256 & -22\% & 26.12  & 23.58
\\ \hline
\end{tabular}
}
\vspace*{.5cm}
\end{table*}

\subsubsection{Transformer performance improvement}

We further observe that limited amounts of in-training factorization seem to improve performance with the transformer model. Reducing the model size by 19\% and 27\% via in-training factorization on the embedding and projection layers improves the BLEU score of the model by 0.3 and 0.2 points, respectively. Additionally, compressing the model by 22\% via in-training factorization across the embedding, projection, feed-forward, and attention layers yields a 0.2 increase in BLEU score. 

We hypothesize that, for the transformer model, in-training factorization serves as a form of regularization to reduce overfitting. As the transformer model is more sensitive to hyperparameter tuning than the LSTM model, as well as showing to be more prone to overfitting, this behavior, while unexpected, is not entirely surprising. 

To test our hypothesis that in-training factorization can increase the performance of the transformer model, we perform the same experiment comparing a transformer model with factorized embeddings of dimension 512 and inner size of 256, with a non-factorized baseline model with embeddings of dimension 512 on two other languages of the IWSLT dataset: Portuguese (79525 training sentences) and Turkish (193734 training sentences). We report our results in \Cref{tab : DEPTTR}. We observe that in all language pairs evaluated, in-training factorization improves the performance of the transformer model. This evidences that the performance gains observed with in-training factorization generalize across languages.

\begin{table*}
\centering
\caption{ \label{tab : DEPTTR} Factorized and non-factorized transformer performance by language. The non-factorized model uses an embedding dimension of 512, and the factorized model uses an inner size of 256.}
\vspace{0.5cm}
\begin{tabular} {|l|l|l|l|}
\hline
 & BLEU with in-training factorization & BLEU without factorization
\\ \Xhline{3\arrayrulewidth}
German & \textbf{26.22} & 25.95
\\ \hline
Portuguese & \textbf{22.63} & 21.56
\\ \hline
Turkish & \textbf{12.05} & 11.95
\\ \hline
\end{tabular}
\end{table*}

\begin{table*}
\centering
\caption{ \label{tab : transformer_by_language} Transformer model parameters for each language.}
\vspace{0.5cm}
\resizebox{\textwidth}{!}{
\begin{tabular} {|l|l|l|l|l|l|}
\hline
 & \textbf{Layers} & \textbf{Embedding dim.} & \textbf{Feed-forward dim.} & \textbf{Attention dim.} & \textbf{Total batch size}
\\ \Xhline{3\arrayrulewidth}
German & 3 & 512 & 1024 & 512 & 128
\\ \hline
Portuguese & 4 & 512 & 512 & 512 & 96
\\ \hline
Turkish & 5 & 512 & 512 & 512 & 96
\\ \hline
\end{tabular}
}
\end{table*}

\subsubsection{Influence on training time}

\begin{figure}
    \centering
    \includegraphics[width=0.6\textwidth]{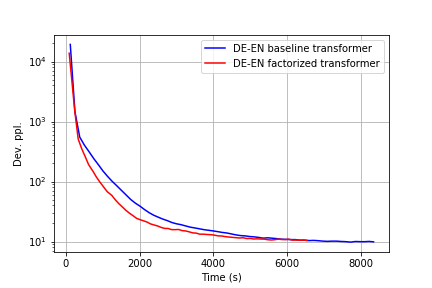}
    \caption{Time evolution of validation perplexity for baseline and in-training factorized transformers with the same GPU memory usage. The factorized model has half the parameters and double the batch size.}
    \label{fig : training-time}
\end{figure}

We additionally examine the impact of in-training factorization on training time for the transformer model. Compressing the model size allows for larger batches to to be stored in the same amount of memory. Thus, if we compress a model by greater than 50\%, then we should be able to double the batch size. We compare the time to convergence for our baseline non-factorized transformer model with an embedding dimension of 512 with an in-training factorized model. The factorized model has a 51\% reduction in model size, and thus we can double the batch size than that of the non-factorized model. The results of this experiment are provided in \Cref{fig : training-time}. We observe that the factorized transformer model converges about 20\% faster than the baseline model. This confirms the findings from \cite{DBLP:journals/corr/abs-1804-00247} that increasing the batch size can decrease training time. 

\subsection{Post-training compression}\label{sec : disc-ps}

\subsubsection{Post-training factorization}

After performing the in-training experiments, we compare these results with traditional post-training factorization methods. To select rank values for post-training factorization, we compute the singular values of the weight matrices of the transformer model in its different layer functions (embedding and projection, attention, and feed-forward) to determine which are easiest to factorize. We find that the singular values of the weight matrices have very different structures. Those of the embedding and projection layers are dominated by the biggest singular value, around 5 times bigger than the second biggest. On the contrary, the attention and feed-forward matrices are more balanced. As the main difference between the two kinds of matrices is that there is no bias vector added after the embeddings, our hypothesis is that this large singular value is a form of learned bias that skews the parameters heavily.

We consider a singular value relevant if it is bigger than 10\% of the dominant one. As the embedding and projection layers are dominated by one value that is much bigger than the rest, we find that no more than 1\% of values are within those bounds, which is coherent with the better results we obtain on those layers. The attention and feed-forward layers, in contrast, seem well suited to halving their rank (for both, around 35\% to 60\% of the singular values are relevant according to our heuristic). 

We use a rank of 256 (half of the singular values) and attempt to factorize the whole model. We experiment with a variety of compression schemes on the transformer model with those values, and report the results in \Cref{tab : transformer post-training}. We observe that, similar to the in-training factorization, the post-training factorization also improves the performance of the baseline transformer model. This provides more evidence that matrix factorization can function as a regularizer for models that are prone to overfitting. We additionally observe that the performance of the post-training factorization seems marginally better than the performance of the in-training factorization, although this might be due to additional training.

\subsubsection{Post-training factorization and pruning}

Lastly, as pruning and factorization have similar goals, we attempt to combine their gains by pruning a post-training factorized model (\Cref{tab : transformer post-training}). We observe that pruning an already factorized model decreases BLEU score by nearly 5 points from the score of the factorized model (from 26.38 to 21.48). This strongly suggests that the parameter reductions for both methods are, at least partly, redundant.

\begin{table*}
\centering
\caption{ \label{tab : transformer post-training}Comparison of in-training and post-training factorization for the transformer model on the German test set. ``In-training (+attention)" refers to in-training factorization applied to the embedding, projection, feed-forward, and attention layers. ``Post-training" refers to post-training factorization applied to all weights. ``Post-training then pruned" refers to pruning occurring after post-training factorization.}
\vspace{0.5cm}
\begin{tabular}{|l|l|l|l|}
\hline
\textbf{Compression method} & \textbf{Size reduction} & \textbf{BLEU} & \textbf{BLEU non-factorized}
\\ \Xhline{3\arrayrulewidth}
None, embedding=512 (baseline) & 0\% & - & 25.95
\\ \Xhline{2\arrayrulewidth}
In-training (+attention) & -22\% & 26.12 & -
\\ \hline
Post-training & -22\% & 26.44 & -
\\ \hline
Post-training then pruned & -46.8\% & 21.48 & -
\\ \hline
\end{tabular}
\end{table*}

\section{Related work} \label{sec : related}

\subsection{Neural Machine Translation}

Neural networks have taken an increasing share of sequence-to-sequence models in the past few years. \citet{Bahdanau14} first use attentional encoder-decoders for machine translation, establishing a new baseline for many incremental additions \citep{luong2015, liu16, gu17}. A few years later \cite{vaswani2017} introduced transformer networks using self attention as an efficient alternative for machine translation tasks. State-of-the-art versions of these models are usually applied on subword units obtained through methods like Byte-Pair Encoding \citep{britz17}.

\subsection{Model compression}

In order to save on parameters, \citet{see2016} propose to use \textit{parameter pruning} to reduce model size by replacing the n\% smallest elements in the network with 0. They show that they can reduce the number of weights in a LSTM encoder-decoder by 40\% without significant drop in performance -- which we reproduced. \citet{DBLP:journals/corr/MurrayC15} is the equivalent of our training time compression for pruning, learning which weights are necessary as training goes on and pruning them accordingly.  Other compression methods involve \textit{quantization} at inference \citep{Wu2016,devlin2017, quinn2018} or in training \citep{Micikevicius2017} ; or for example involve \textit{model distillation} \citep{Kim2016}.

\subsection{Parameter matrix factorization}

\textit{Matrix factorization} is a wide family of algorithms \citep{DBLP:journals/corr/Yang15b}. Accordingly, it has been used in very diverse ways to help compress neural networks. The most common is to use a low-rank approximation to compress Convolutional Neural Networks without loss of performance, as \citet{6638949} do for speech recognition and \citet{DBLP:journals/corr/KimPYCYS15} for computer vision. \citet{Prabhavalkar16} use low-rank factorization on inner layers of the LSTM for speech recognition, as does \citet{inproceedings} in a constrained version. The work in \citet{Stahlberg17} is closer to ours, as they factorize trained embedding matrices in encoder-decoders with SVD. 

However, all of these uses are post-training compression techniques with a simple application of SVD. We focus on in-training factorization, and compare that approach with the existing post-training factorization paradigm. The closest work that we know of is in \citet{Kuchaiev2017FactorizationTF}, which factorizes the LSTM cell matrices during training in a highly optimized framework; however, they do so at the cost of an increase of perplexity. On the other hand, we focus on embedding and attention matrices, for which this work hasn't been performed before to the best of our knowledge, with no cost, or even small gains, in performance.

\section{Future work}

A limitation of our paper is that it does not explore how to select the hidden sizes for in-training factorized models. Exploring a priori strategies for selecting hidden sizes for in-training factorization could be a valuable extension of our work. Another extension of our work could include analyzing why transformer models consistently improve performance when learned with in-training factorization. It may also be constructive to examine whether our results generalize to larger machine translation corpora. Other potential areas of research could include studying the impact of in-training factorization on other natural language processing tasks, including classification tasks and word2vec embeddings. Finally, a robust comparison between in-training factorization and post-training factorization on model performance would likely be productive. 


\section{Conclusion}

We have introduced a method to use matrix factorization at training time to reduce the parameter footprint of neural machine translation models. We compare in-training factorization to existing post-hoc parameter reduction methods, including parameter pruning and post-training factorization. We find that using factorization comes with significant gains in both final performance and number of required parameters. Lastly, we demonstrate the effectiveness of our in-training factorization technique to learn a model with fewer parameters, improved accuracy, and decreased training time.

\bibliography{iclr2020_conference}
\bibliographystyle{iclr2020_conference}

\end{document}